\documentclass[conference]{IEEEtran}
\IEEEoverridecommandlockouts

\usepackage{cite}
\usepackage{amsmath,amssymb,amsfonts}
\usepackage{algorithmic}
\usepackage{graphicx}
\usepackage{textcomp}
\usepackage{xcolor}

\def\BibTeX{{\rm B\kern-.05em{\sc i\kern-.025em b}\kern-.08em
    T\kern-.1667em\lower.7ex\hbox{E}\kern-.125emX}}
\begin{document}

\makeatletter
\newcommand{\newlineauthors}{
  \end{@IEEEauthorhalign}\hfill\mbox{}\par
  \mbox{}\hfill\begin{@IEEEauthorhalign}
}
\makeatother

\title{An Intelligent Approach to Detecting Novel Fault Classes for Centrifugal Pumps Based on Deep CNNs and Unsupervised Methods}
\author{\IEEEauthorblockN{Mahdi Abdollah Chalaki}
\IEEEauthorblockA{\textit{School of Mechanical Engineering}\\
\textit{University of Tehran} \\
Tehran, Iran \\
mahdichalaki@ut.ac.ir}
\and
\IEEEauthorblockN{Daniyal Maroufi}
\IEEEauthorblockA{\textit{School of Mechanical Engineering}\\
\textit{University of Tehran} \\
Tehran, Iran \\
daniyalmarofi@ut.ac.ir}
\and
\IEEEauthorblockN{Mahdi Robati}
\IEEEauthorblockA{\textit{School of Mechanical Engineering}\\
\textit{University of Tehran} \\
Tehran, Iran \\
mahdi.robati@ut.ac.ir}
\newlineauthors
\IEEEauthorblockN{Mohammad Javad Karimi}
\IEEEauthorblockA{\textit{School of Mechanical Engineering}\\
\textit{University of Tehran} \\
Tehran, Iran \\
jvdkarimi@ut.ac.ir}
\and
\IEEEauthorblockN{Ali Sadighi}
\IEEEauthorblockA{\textit{School of Mechanical Engineering}\\
\textit{University of Tehran} \\
Tehran, Iran \\
asadighi@ut.ac.ir}
}

\maketitle

\begin{abstract}
Despite the recent success in data-driven fault diagnosis of rotating machines, there are still remaining challenges in this field. Among the issues to be addressed, is the lack of information about variety of faults the system may encounter in the field. In this paper, we assume a partial knowledge of the system faults and use the corresponding data to train a convolutional neural network. A combination of t-SNE method and clustering techniques is then employed to detect novel faults. Upon detection, the network is augmented using the new data. Finally, a test setup is used to validate this two-stage methodology on a centrifugal pump and experimental results show high accuracy in detecting novel faults.
\end{abstract}


\section{Introduction}
Diagnosis of rotating machinery faults is essential since it leads to reduction of unplanned downtime and \mbox{maintenance} costs and improves reliability and safety of operations. \mbox{Centrifugal} pumps are one of the most-commonly used equipment in a wide range of industries. Researchers have done \mbox{extensive} research on implementing online condition \mbox{monitoring} methods in order to detect faults effectively.

These methods comprise data-driven approaches such as Multi-layer Perceptron (MLP) \cite{hajnayeb2011application}, Random forests \cite{yang2008random}, \mbox{Support} Vector Machines\cite{santos2015svm}, etc., which were used alongside extracted statistical features and signal-processing techniques. Recently, various deep learning approaches have shown remarkable results on classification tasks \cite{li2020intelligent,zhao1612deep}. These \mbox{algorithms} mainly use labeled raw time-domain sensor signals as input and learn the pattern within this data to extract main features and classify the faults.

In spite of these impressive results, there are remaining challenges that limit the application of these methods in real-world plants. The utilization of most existing methods is limited by some open issues such as domain adaptation problems, imbalanced datasets and missing labels. Different approaches are taken to address each of these issues in recent studies \cite{li2020deep,li2020intelligent,wang2020missing}. However, the problem of novel faults was not considered by any of them.

To develop statistical models capable of classifying \mbox{machine} faults, sufficient amount of labeled data is required. Such a data is not readily available in industry settings. \mbox{Alternatively}, labeled data can be generated in the laboratory by \mbox{introducing} various faults and using them for model development. \mbox{However}, novel faults not seen by the model will result in misclassifications and poor performance of the monitoring system. Anomaly detection algorithms, on the other hand, can distinguish healthy state of the system from any anomalous behavior \cite{jamimoghaddam2020esa,masoumauto}. Nevertheless, different faults cannot be discerned using these methods. In this paper, we propose a new approach that combines a supervised convolutional neural \mbox{network} (CNN) model with an unsupervised clustering network. Consequently, not only does the system detect previously observed faults, but it also groups new faults into new clusters and continuously monitors their types.

\section{An Introduction to CNNs}

\begin{figure*}[t]
\centering
\includegraphics[width=\textwidth]{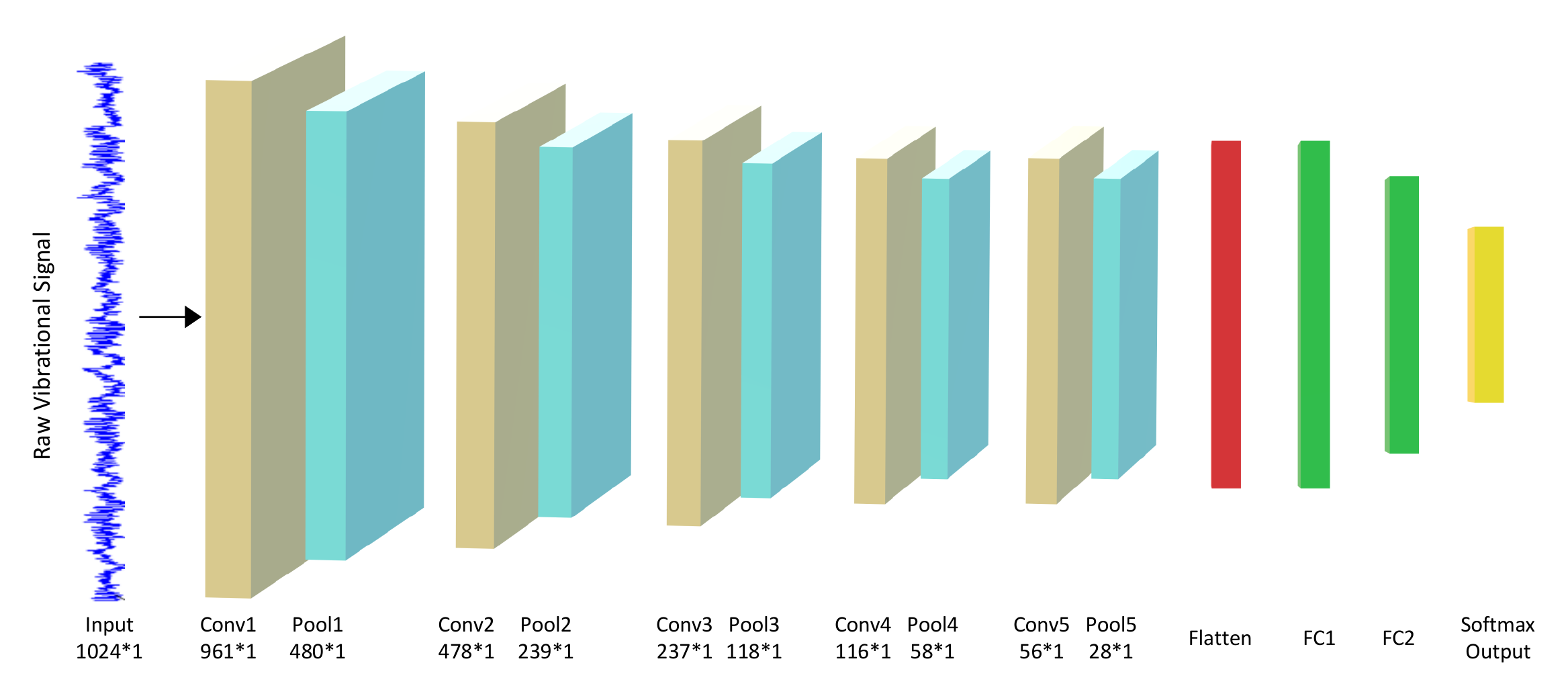}
\caption{Architecture of the WDCNN model. "Conv" stands for a Convolutional layer, "Pool" for a Max-Pooling layer, and "FC" for a Fully-Connected layer.}
\label{fig:dcnn}
\end{figure*}

Convolutional neural networks (CNNs) are capable of \mbox{learning} high-level features from input data, which allows them to provide accurate classification. There are three types of layers in convolutional neural networks: convolutional, pooling, and fully connected. In most classification and regression \mbox{applications}, the last layer is usually a fully connected layer.  A CNN is based on three ideas: shared weights, local receptive fields, and pooling. In contrast to fully connected neural networks, in a CNN only a small localized area of the input data is connected to each node of a convolutional layer. Thus, it reduces the number of parameters of the network and makes it easier to train. 

In a CNN architecture, a block consists of a convolution layer and a pooling layer. A number of these blocks followed by fully-connected layers form a deep network that is used for feature extraction and classification. Finally, a softmax function predicts the probability of each class \cite{gu2018recent,hinton2012improving}. Mathematically, it is determined by:

\begin{equation}
    y = f(W a+b_c) 
    \label{eq:softmax}
\end{equation}
where $y$ denotes the predicted labels; $W$ denotes the weight matrix between $a$, neurons in fully-connected layer and \mbox{connected} nodes in the output layer; $b_c$ represents the bias, and $f$ is the softmax activation function. The training of the CNN is done by minimizing the cross-entropy loss, which is calculated as:

\begin{equation}
    H(r,p) = - \sum_{i} r_i log(p_i)
    \label{eq:softmax2}
\end{equation}

According to the true label, $r$ is either 0 or 1, and $p$ is the output probability. The error between predicted probabilities and true labels can be measured by cross-entropy loss. Finally, weights will be updated after each iteration using a backpropagation algorithm based on gradient descent methods.

\section{Proposed Methodology}

As mentioned before, although deep learning methods have shown great success in classifying the faults that they have been trained on, they fail to detect new types of faults. Here, a new network is proposed to address this problem.

This proposed network consists of two stages: In the first one, a Deep Convolutional Neural Networks with Wide First-layer Kernels (WDCNN) \cite{zhang2017new} is used to extract features and classify faults, and in the second stage faults are examined for their novelness. If a novel fault is encountered, the monitoring system could automatically detect it, make a new label and modify the current network to include the new class.

\subsection{Stage 1}  \label{sec:stage1}

In the first stage, a network capable of classifying the faults of the system is built. The structure of this neural network is heavily borrowed from \cite{zhang2017new}. Raw vibration signals are used as inputs to this network. In the first convolutional layer, feature extraction and high-frequency noise suppression are conducted by using wide kernels. The network structure is visualized in Fig. \ref{fig:dcnn}.

Batch normalization\cite{ioffe2015batch} and Dropout\cite{dropout} layers after each convolutional layer are employed to decrease the training process duration and prevent overfitting. Additionally, categorical crossentropy is used to measure loss function and the Adam optimizer\cite{kingma2017adam} is employed instead of the classic SGD to train the model.

After a successful training stage, this convolutional neural network can classify faults into a certain number of classes determined by our training dataset. As shown in Fig. \ref{fig:dcnn}, the last fully connected layer is made up of 64 neurons that can extract the features of each sample. These features play an essential role in the second stage.

\subsection{Stage 2} \label{sec:stage2}

In this stage, the dimensions of the features extracted in the previous stage are reduced by applying the t-SNE method \cite{van2008visualizing}. In our case, the features can be described as a 64-dimensional vector, whose dimension could be reduced to two. We then use these two-dimensional features to form clusters. As it is shown in Fig. \ref{fig:tsne_6class}, for 600 training samples (from 6 different labels), each class forms a unique cluster after this process.

\begin{figure}[t]
\centerline{\includegraphics[width=\linewidth]{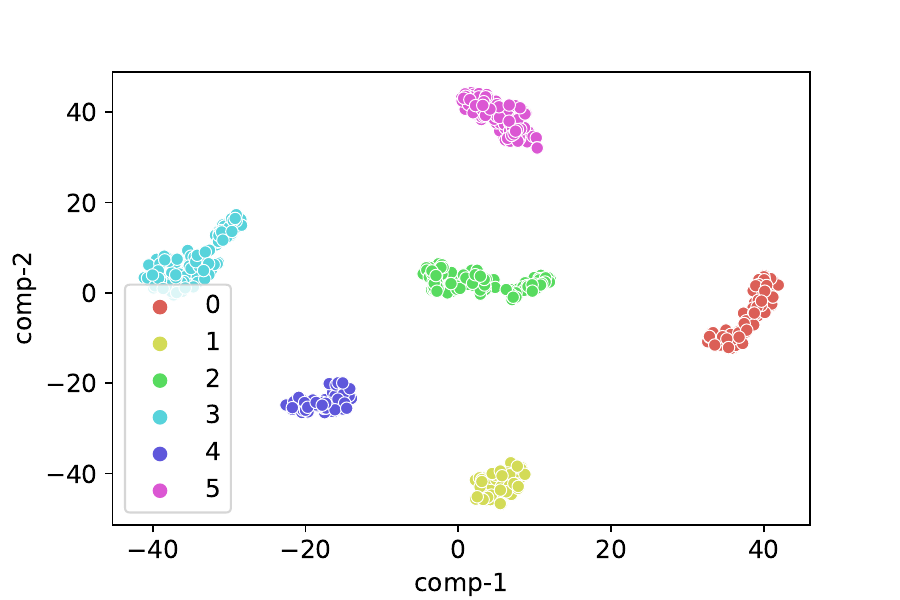}}
\caption{Feature visualization of the last hidden FC layer of WDCNN network via t-SNE.}
\label{fig:tsne_6class}
\end{figure}

If the machine faces a faulty condition of any kind, the system will start collecting 100 new samples. By averaging over the network predictions for these new samples, it can be concluded that there exists a problem. In order to determine if these samples form a new cluster, they will be clustered with previous samples. K-means algorithm is used in this process to cluster the samples \cite{1056489}.

K-means clustering is an unsupervised learning technique that minimizes intra-cluster variances to divide $n$ unlabeled samples into a pre-determined number of clusters with the nearest mean.

The goal here is to determine if the newly collected samples create a distinct cluster or not. Thus, firstly we have to try different number of clusters to partition samples into, and then use an evaluation method to find the optimal number of clusters.

External, relative, and internal techniques are usually used for cluster validation. Using the internal criteria, which is an unsupervised way to estimate the quality of a \mbox{clustering} \mbox{solution}, we measure how compacted and separated the \mbox{clusters} are. This is a practical method that can help us to determine how many clusters exist.

There are several internal criteria indices suggested in the literature including Davies–Bouldin index \cite{4766909}, Dunn’s Index\cite{dunn}, The Silhouette index \cite{ROUSSEEUW198753}  and indices based on Sum of Squared estimate of Errors (SSE). The proposed method uses the latest one in combination with an inflection point selection method to find the knee-point in the "Inertia - Number of clusters" diagram. Inertia is defined here as the squared distance of the samples to their nearest cluster center \cite{lloyd1982least}. This method is robust to the absolute value of the data points (Unlike other methods in which a limit should be set to find out if two samples are in the same neighborhood).

Knee-point analysis indicates whether a new cluster has appeared in the data. Should this occur, the system begins collecting more data. As training a new model that includes a new label requires a balanced dataset, this data collection step is necessary. After the data collection stage, we turn back to Stage 1. The next step is to modify the softmax layer of our old model and add a neuron to its neurons. After this modification, the model will be trained again with the new dataset.

\begin{table}[!b]
\renewcommand{\arraystretch}{1.3}
\caption{Description of centrifugal pump dataset}
\label{table1}
\centering
\begin{tabular}[width=\linewidth]{|c|c|c|c|c|}
\hline
Type          & Train & Test & Flow range(L/min) & Label \\
\hline
Healthy       & 1460           & 360        & 170 - 220           & 0     \\
\hline
Low flow rate & 1120           & 280        & 80 - 120            & 1     \\
\hline
Cavitation    & 1440           & 360        & 120 - 170           & 2     \\
\hline
Major defect  & 1440           & 360        & 150 - 200           & 3     \\
\hline
Minor defect  & 1040           & 360        & 130 - 160           & 4     \\
\hline
Crack         & 1280           & 320        & 150 - 200           & 5    \\
\hline
\end{tabular}
\end{table}

\section{Experimental Investigation} \label{sec:exp}

\subsection{Experimental Setup} \label{sec:setup}

To evaluate the proposed method, an experimental setup was developed as illustrated in Fig. \ref{fig:closedloop}. This system consists of a reservoir, a centrifugal pump, a rotameter, and two valves that control water flow. A 1-hp, single-phase induction motor drives the water pump.

Two valves are located at the suction and discharge sides of the pump. Cavitation is staged by changing the inlet valve position, and flow rate is controlled by adjusting the outlet valve.

\begin{figure}[t!]
\centerline{\includegraphics[width=\linewidth]{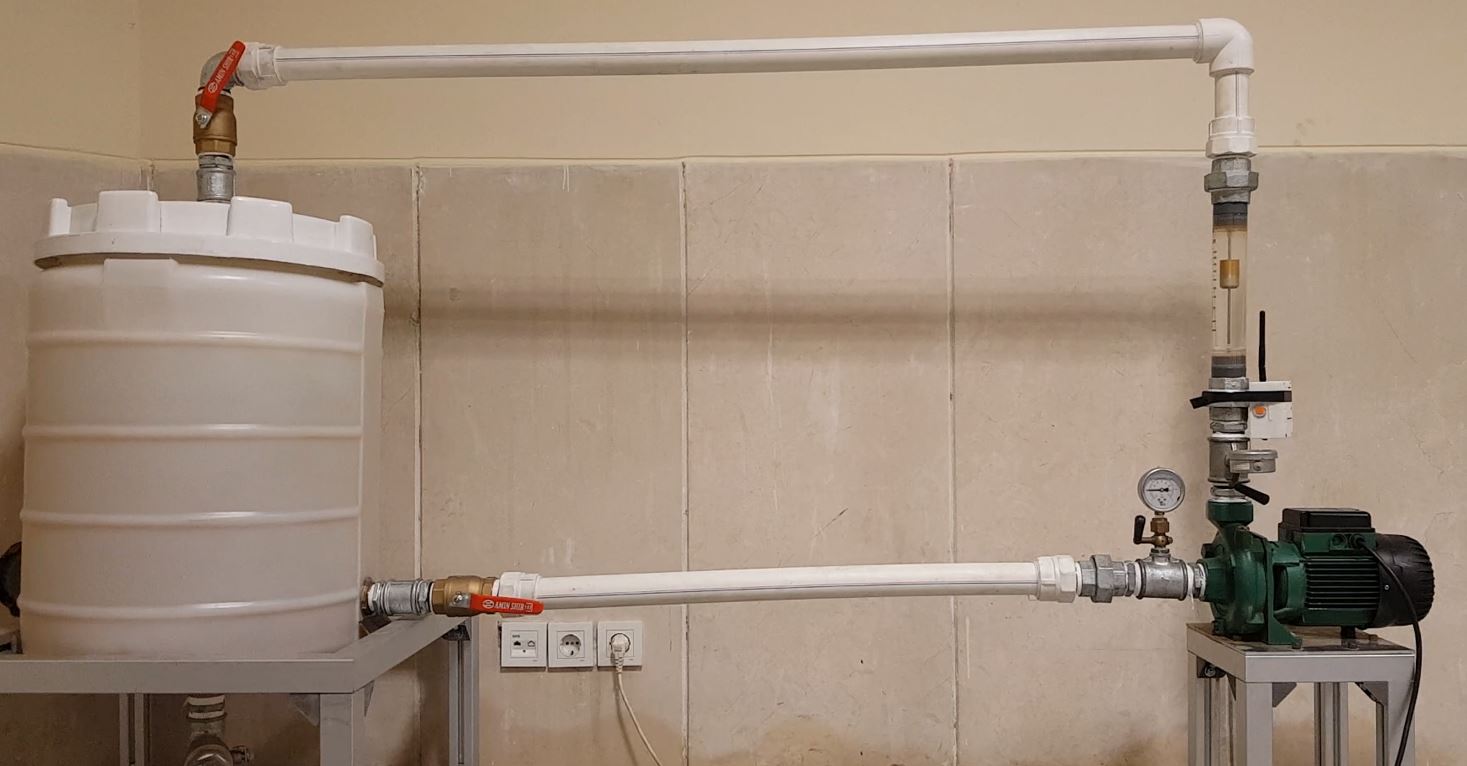}}
\caption{Experimental setup.}
\label{fig:closedloop}
\end{figure}

The vibration signal is sampled at 1.6 kHz using a LIS3DSH accelerometer module that is mounted on the pump's discharge side. This module is a three-axis accelerometer which includes an 800 Hz anti-aliasing filter. Due to the limited memory of our microcontroller, this model only uses the data from one of the axes. In addition, using data from a single axis means fewer parameters and hence, less computational resources.

The accelerometer module is connected to an STM32F4-Discovery microcontroller using Serial Peripheral Interface (SPI) protocol. Sampled data is first stored on the microcontroller memory and then transmitted to the server through two XBee modules, one installed on the main device which is mounted on the pump, and the other in the monitoring station. XBee modules use ZigBee protocol to communicate with each other, and UART protocol to interface with the microcontroller on the device and the Raspberry Pi board on the user end. This process is illustrated in Fig. \ref{fig:setup}.

\begin{figure}[t!]
\centerline{\includegraphics[width=\linewidth]{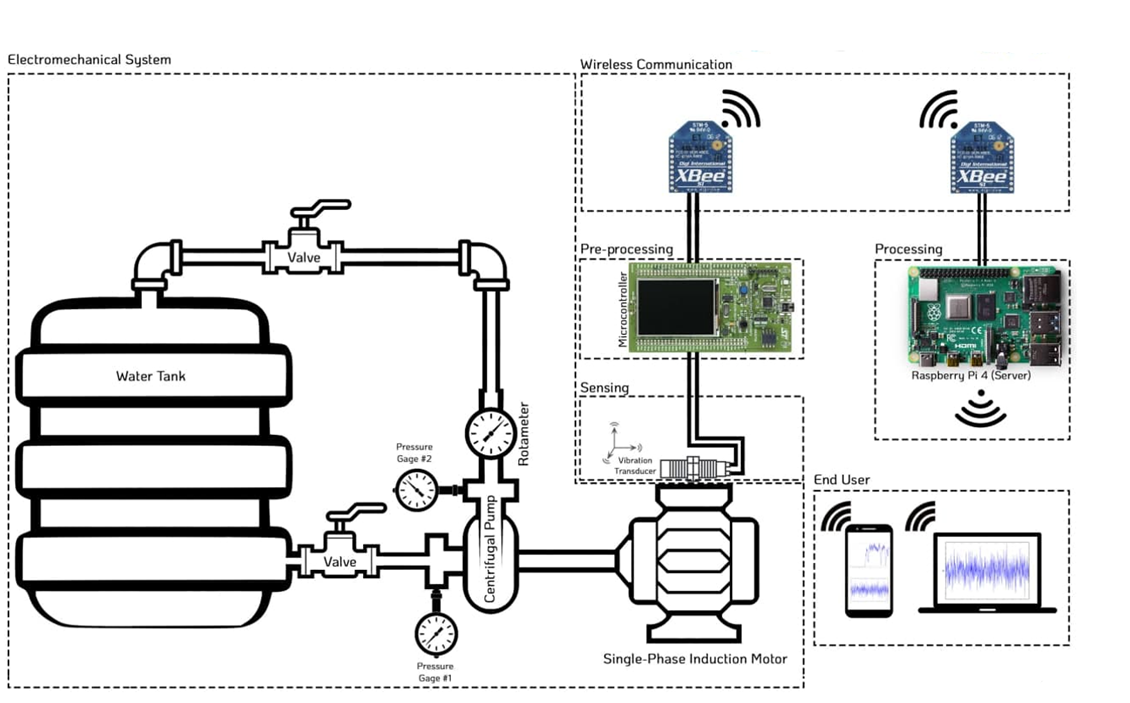}}
\caption{Schematic picture of the Experimental setup and its communication with the user-end part.}
\label{fig:setup}
\end{figure}

The Raspberry Pi board first stores raw data in an InfluxDB database, and then does the pre-processing task. In this task, the data is reshaped to our favorable form, get normalized and then stored in the database again. At last, this pre-processed data is sent to the CNN model for continuous evaluation.

Finally, the pre-processed data batches and their \mbox{corresponding} predicted labels are shown in a Grafana dashboard for online monitoring purposes of the end-user. Fig. \ref{fig:pipeline} schematically shows the software architecture.

\begin{figure}[t!]
\centerline{\includegraphics[width=\linewidth]{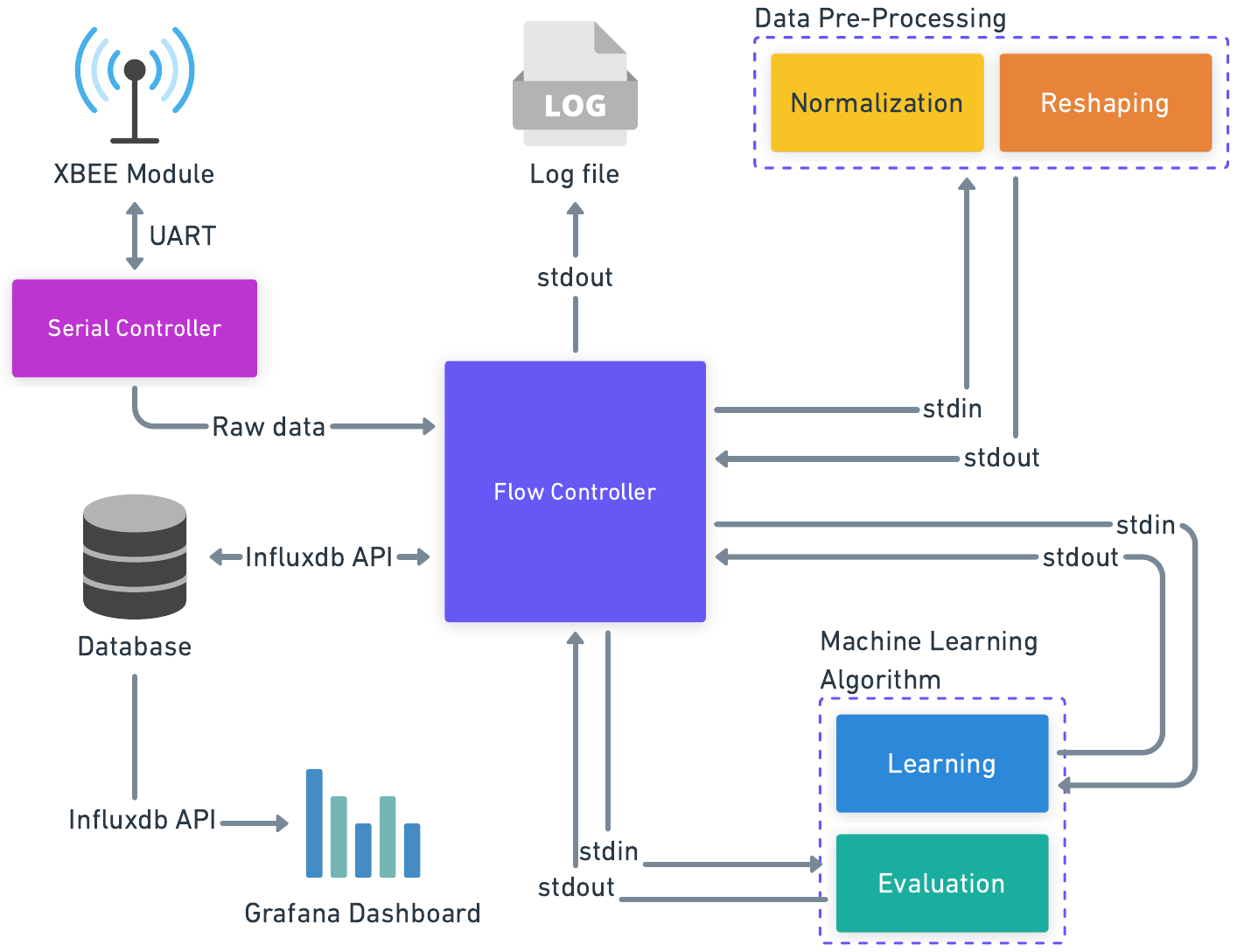}}
\caption{Software architecture.}
\label{fig:pipeline}
\end{figure}

\begin{figure}[t!]
\centerline{\includegraphics[width=0.9\linewidth]{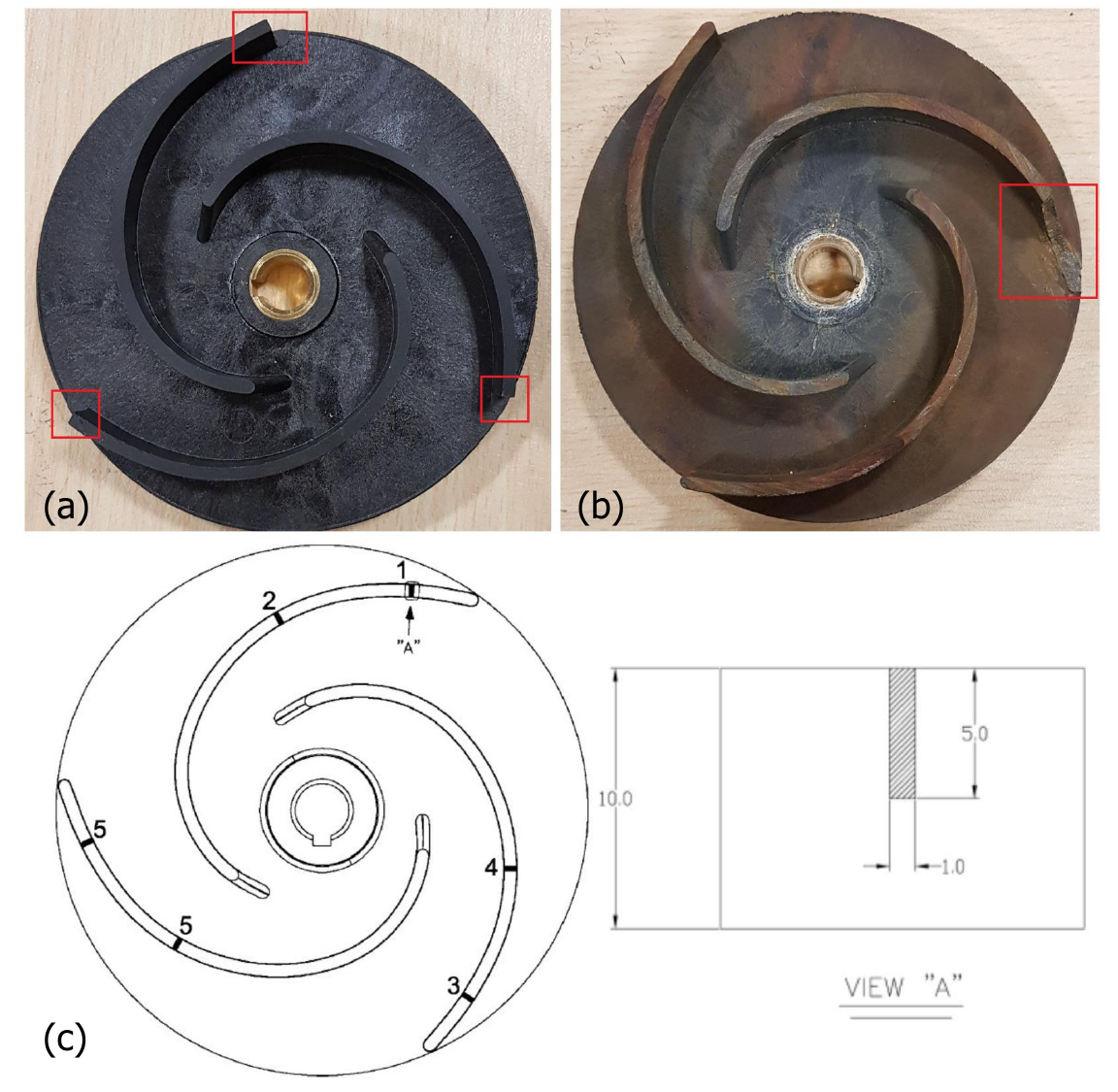}}
\caption{(a) vane tip minor defect
(b) vane tip major defect
(c) impeller crack fault \cite{jamimoghaddam2020esa}.}
\label{fig:defects}
\end{figure}

\subsection{Dataset Description} \label{sec:dataset}

Our dataset consists of 6 types of health conditions; a healthy condition and five different faulty types. Each sample contains 1024 data points. Additional details are shown in Table \ref{table1}.

Fig. \ref{fig:defects} provides a better understanding of the impeller-related faults by illustrating vane tip defects (corresponding to the third and fourth labels) and impeller crack fault (corresponding to the fifth label).

In addition, low flow rate and cavitation conditions are adjusted by using the valves that were previously mentioned.

\section{Results}

In the first stage of training, we just used the first five labels to train the model and the sixth label (Crack fault) is left out for evaluation of the proposed method. It is a more realistic situation where we are able to only make a limited number of faults for training. At the end of this phase, our network can successfully detect samples from each class. Fig. \ref{fig:cm_5class} depicts the confusion matrix at the end of the first stage.

\begin{figure}[t!]
\centerline{\includegraphics[width=0.9\linewidth]{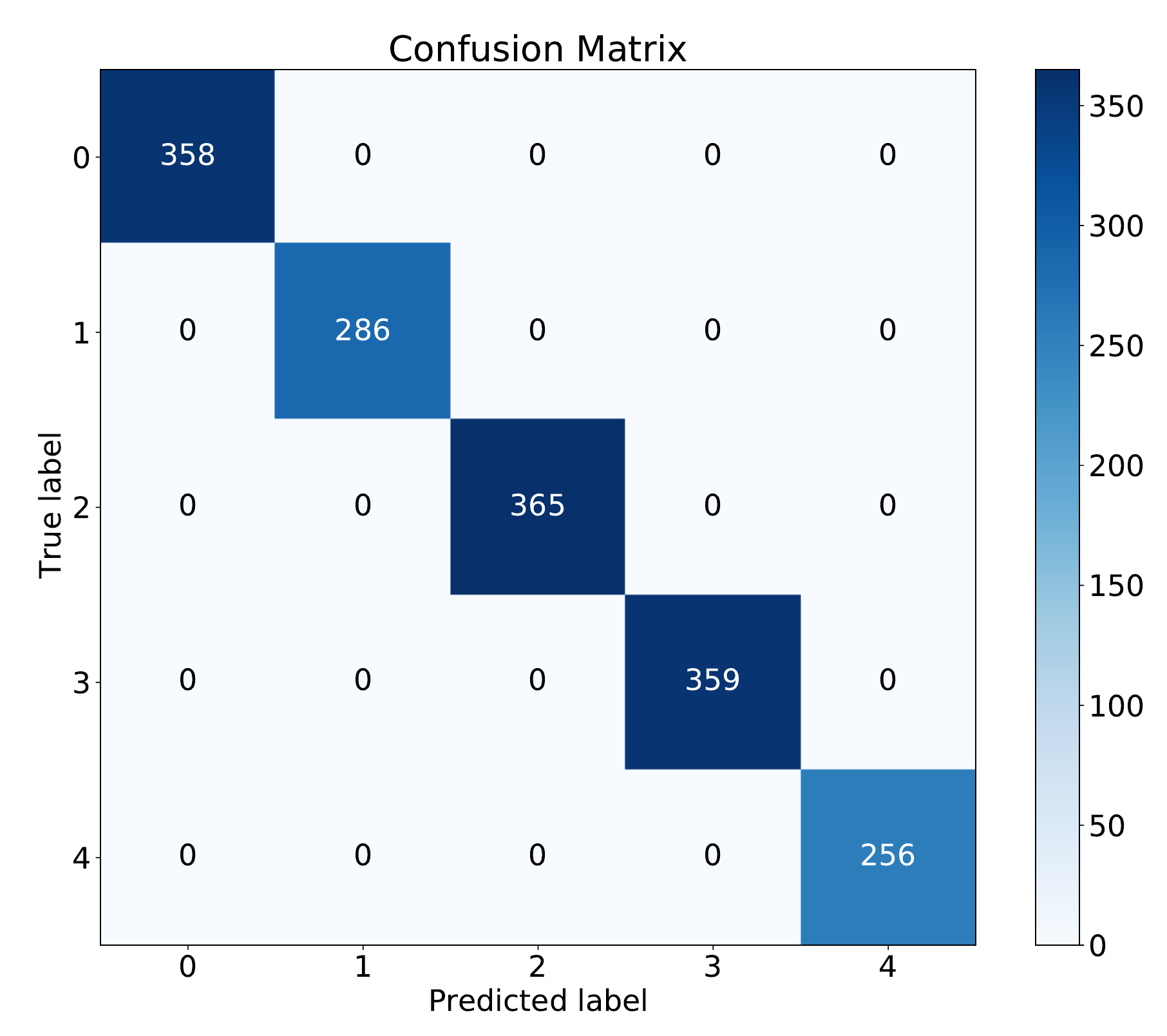}}
\caption{Results of the WDCNN network with 5 classes.}
\label{fig:cm_5class}
\end{figure}

\begin{figure}[t!]
\centerline{\includegraphics[width=0.85\linewidth]{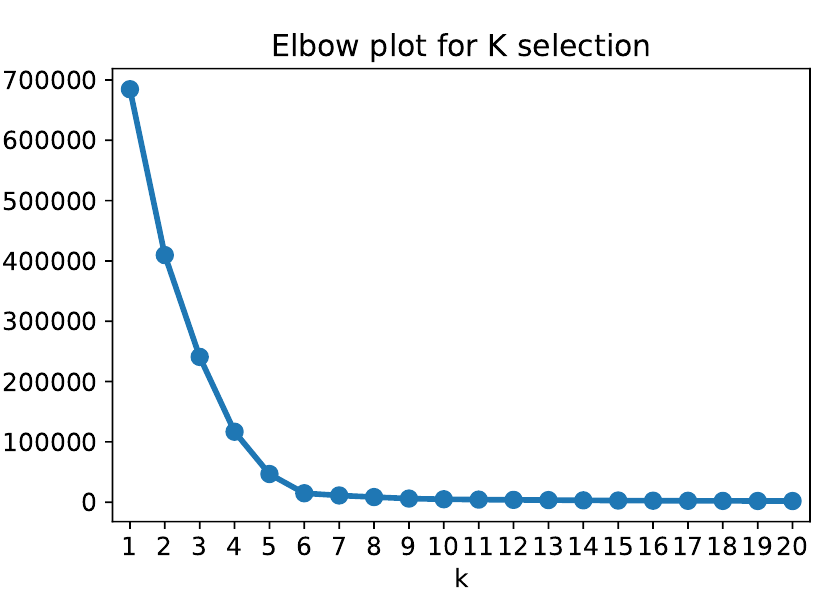}}
\caption{Sum of Squared estimate of Errors (SSE) for different number of clusters.}
\label{sse}
\end{figure}

\begin{figure}[t!]
\centerline{\includegraphics[width=0.85\linewidth]{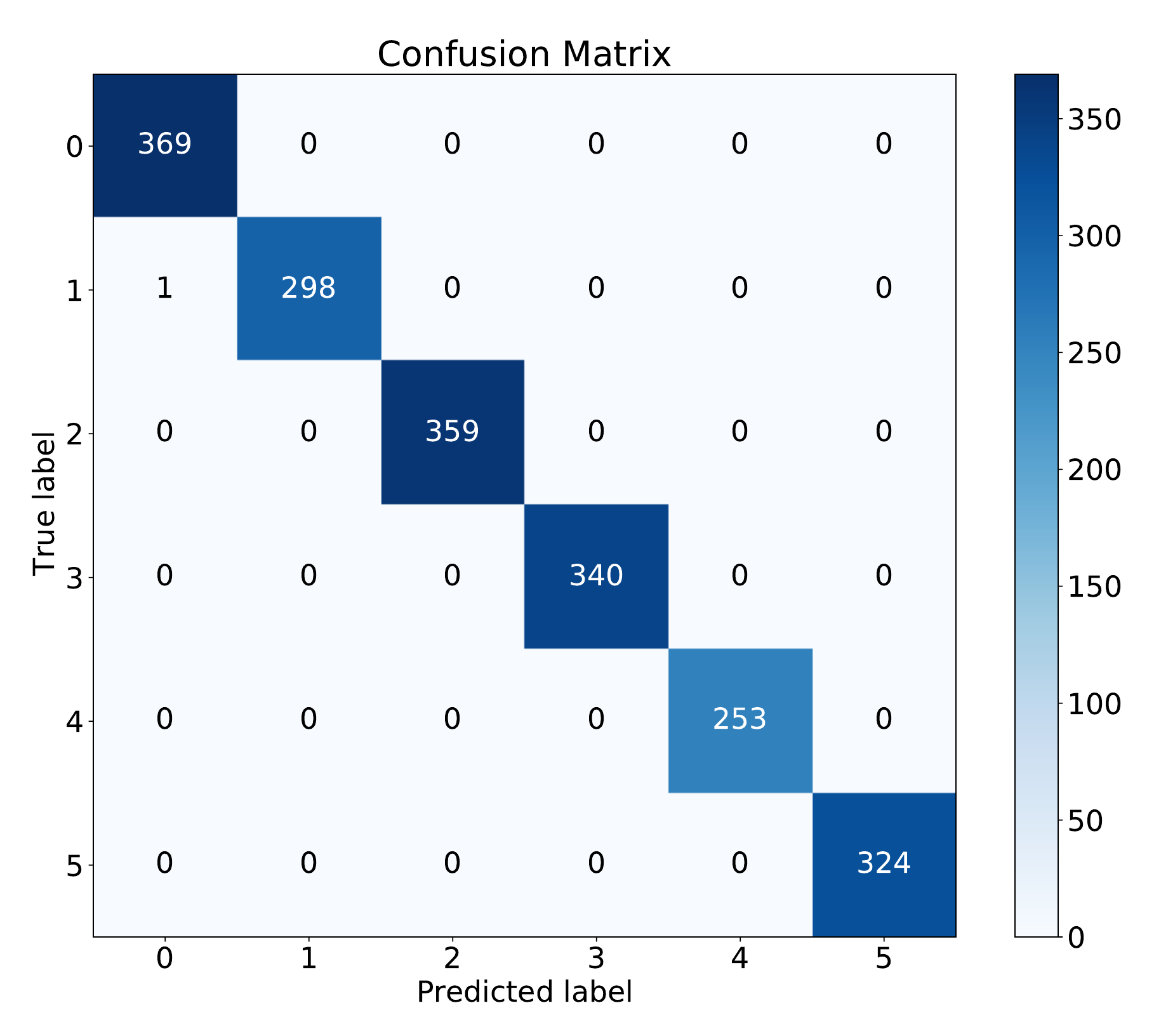}}
\caption{Results of the re-trained network with 6 classes.}
\label{cm_6class}
\end{figure}

In the second stage, the crack fault is introduced to the model. Averaging over the predictions of the last 100 samples shows that the pump is not in a healthy condition. To determine if this fault belongs to one of the existing labels, we perform the second stage explained in Section \ref{sec:exp}. These 100 new faulty samples are combined with 500 random samples from our initial dataset. The output of applying the t-SNE on these 600 samples is shown in Fig. \ref{fig:tsne_6class}.

To automatically find the number of clusters, K-means algorithm is applied with different numbers of clusters ranging from 1 to 20. This helps us to find the knee-point of this diagram, as well as the number of clusters. In our case, the procedure indicates that there are six clusters now (Fig. \ref{sse}).

Having a new cluster, sufficient data is needed to train a new model. In an actual situation, this means collecting data for a determined period of time. In our experiment, we used the data from Class 5. A slight modification was also necessary (changing the softmax layer so it will have six outputs). To evaluate the proposed method, we used K-fold Cross-Validation with five folds. The results of this test on our new model are shown in Table \ref{table2}.

As shown in Table \ref{table2}, the proposed method could \mbox{successfully} address the problem of classifying new emerging faults in a centrifugal pump. Fig. \ref{cm_6class} illustrates confusion matrix of the re-trained model.

\begin{table}[!t]
\renewcommand{\arraystretch}{1.3}
\caption{Test accuracy of the re-trained model measured with K-fold Cross-Validation}
\label{table2}
\centering
\begin{tabular}[width=\linewidth]{|c|c|}
\hline
Fold          & Test accuracy(\%) \\ \hline
1          & 99.948 \\ \hline
2          & 100.0 \\ \hline
3          & 100.0 \\ \hline
4          & 100.0 \\ \hline
5          & 100.0 \\ \hline
Average          & 99.989 $\pm$ 0.021 \\ \hline
\end{tabular}
\end{table}

\section{Conclusion}

The paper presented a new two-stage methodology to fault detection of rotating machines. In the first stage, prior \mbox{information} about the system faults was used to develop a CNN capable of classifying defects. The second stage comprised t-SNE and unsupervised clustering algorithms to detect novel fault classes and augment the existing CNN accordingly. A hydraulic test rig was used to implement and show the performance of the proposed method where high accuracy in detecting novel faults was obtained.

\bibliographystyle{IEEEtran}
\bibliography{mybibfile}

\begin{thebibliography}{10}
\providecommand{\url}[1]{#1}
\csname url@samestyle\endcsname
\providecommand{\newblock}{\relax}
\providecommand{\bibinfo}[2]{#2}
\providecommand{\BIBentrySTDinterwordspacing}{\spaceskip=0pt\relax}
\providecommand{\BIBentryALTinterwordstretchfactor}{4}
\providecommand{\BIBentryALTinterwordspacing}{\spaceskip=\fontdimen2\font plus
\BIBentryALTinterwordstretchfactor\fontdimen3\font minus \fontdimen4\font\relax}
\providecommand{\BIBforeignlanguage}[2]{{%
\expandafter\ifx\csname l@#1\endcsname\relax
\typeout{** WARNING: IEEEtran.bst: No hyphenation pattern has been}%
\typeout{** loaded for the language `#1'. Using the pattern for}%
\typeout{** the default language instead.}%
\else
\language=\csname l@#1\endcsname
\fi
#2}}
\providecommand{\BIBdecl}{\relax}
\BIBdecl

\bibitem{hajnayeb2011application}
A.~Hajnayeb, A.~Ghasemloonia, S.~Khadem, and M.~H. Moradi, ``Application and comparison of an ann-based feature selection method and the genetic algorithm in gearbox fault diagnosis,'' \emph{Expert systems with Applications}, vol.~38, no.~8, pp. 10\,205--10\,209, 2011.

\bibitem{yang2008random}
B.-S. Yang, X.~Di, and T.~Han, ``Random forests classifier for machine fault diagnosis,'' \emph{Journal of mechanical science and technology}, vol.~22, no.~9, pp. 1716--1725, 2008.

\bibitem{santos2015svm}
P.~Santos, L.~F. Villa, A.~Re{\~n}ones, A.~Bustillo, and J.~Maudes, ``An svm-based solution for fault detection in wind turbines,'' \emph{Sensors}, vol.~15, no.~3, pp. 5627--5648, 2015.

\bibitem{li2020intelligent}
X.~Li, W.~Zhang, Q.~Ding, and J.-Q. Sun, ``Intelligent rotating machinery fault diagnosis based on deep learning using data augmentation,'' \emph{Journal of Intelligent Manufacturing}, vol.~31, no.~2, pp. 433--452, 2020.

\bibitem{zhao1612deep}
R.~Zhao, R.~Yan, Z.~Chen, K.~Mao, P.~Wang, and R.~Gao, ``Deep learning and its applications to machine health monitoring: A survey. arxiv 2016,'' \emph{arXiv preprint arXiv:1612.07640}, 2016.

\bibitem{li2020deep}
X.~Li and W.~Zhang, ``Deep learning-based partial domain adaptation method on intelligent machinery fault diagnostics,'' \emph{IEEE Transactions on Industrial Electronics}, vol.~68, no.~5, pp. 4351--4361, 2020.

\bibitem{wang2020missing}
Q.~Wang, G.~Michau, and O.~Fink, ``Missing-class-robust domain adaptation by unilateral alignment,'' \emph{IEEE Transactions on Industrial Electronics}, vol.~68, no.~1, pp. 663--671, 2020.

\bibitem{jamimoghaddam2020esa}
M.~Jamimoghaddam, A.~Sadighi, and Z.~Araste, ``Esa-based anomaly detection of a centrifugal pump using self-organizing map,'' in \emph{2020 6th Iranian Conference on Signal Processing and Intelligent Systems (ICSPIS)}.\hskip 1em plus 0.5em minus 0.4em\relax IEEE, 2020, pp. 1--5.

\bibitem{masoumauto}
S.~M.~H. Maasoum, A.~Mostafavi, and A.~Sadighi, ``An autoencoder-based algorithm for fault detection of rotating machines, suitable for online learning and standalone applications,'' in \emph{2020 6th Iranian Conference on Signal Processing and Intelligent Systems (ICSPIS)}.\hskip 1em plus 0.5em minus 0.4em\relax IEEE, 2020, pp. 1--6.

\bibitem{gu2018recent}
J.~Gu, Z.~Wang, J.~Kuen, L.~Ma, A.~Shahroudy, B.~Shuai, T.~Liu, X.~Wang, G.~Wang, J.~Cai \emph{et~al.}, ``Recent advances in convolutional neural networks,'' \emph{Pattern Recognition}, vol.~77, pp. 354--377, 2018.

\bibitem{hinton2012improving}
G.~E. Hinton, N.~Srivastava, A.~Krizhevsky, I.~Sutskever, and R.~R. Salakhutdinov, ``Improving neural networks by preventing co-adaptation of feature detectors,'' \emph{arXiv preprint arXiv:1207.0580}, 2012.

\bibitem{zhang2017new}
W.~Zhang, G.~Peng, C.~Li, Y.~Chen, and Z.~Zhang, ``A new deep learning model for fault diagnosis with good anti-noise and domain adaptation ability on raw vibration signals,'' \emph{Sensors}, vol.~17, no.~2, p. 425, 2017.

\bibitem{ioffe2015batch}
S.~Ioffe and C.~Szegedy, ``Batch normalization: Accelerating deep network training by reducing internal covariate shift,'' 2015.

\bibitem{dropout}
N.~Srivastava, G.~Hinton, A.~Krizhevsky, I.~Sutskever, and R.~Salakhutdinov, ``Dropout: A simple way to prevent neural networks from overfitting,'' \emph{J. Mach. Learn. Res.}, vol.~15, no.~1, p. 1929–1958, Jan. 2014.

\bibitem{kingma2017adam}
D.~P. Kingma and J.~Ba, ``Adam: A method for stochastic optimization,'' 2017.

\bibitem{van2008visualizing}
L.~Van~der Maaten and G.~Hinton, ``Visualizing data using t-sne.'' \emph{Journal of machine learning research}, vol.~9, no.~11, 2008.

\bibitem{1056489}
S.~Lloyd, ``Least squares quantization in pcm,'' \emph{IEEE Transactions on Information Theory}, vol.~28, no.~2, pp. 129--137, 1982.

\bibitem{4766909}
D.~L. Davies and D.~W. Bouldin, ``A cluster separation measure,'' \emph{IEEE Transactions on Pattern Analysis and Machine Intelligence}, vol. PAMI-1, no.~2, pp. 224--227, 1979.

\bibitem{dunn}
\BIBentryALTinterwordspacing
J.~C. Dunn†, ``Well-separated clusters and optimal fuzzy partitions,'' \emph{Journal of Cybernetics}, vol.~4, no.~1, pp. 95--104, 1974. [Online]. Available: \url{https://doi.org/10.1080/01969727408546059}
\BIBentrySTDinterwordspacing

\bibitem{ROUSSEEUW198753}
\BIBentryALTinterwordspacing
P.~J. Rousseeuw, ``Silhouettes: A graphical aid to the interpretation and validation of cluster analysis,'' \emph{Journal of Computational and Applied Mathematics}, vol.~20, pp. 53--65, 1987. [Online]. Available: \url{https://www.sciencedirect.com/science/article/pii/0377042787901257}
\BIBentrySTDinterwordspacing

\bibitem{lloyd1982least}
S.~Lloyd, ``Least squares quantization in pcm,'' \emph{IEEE transactions on information theory}, vol.~28, no.~2, pp. 129--137, 1982.

\end{thebibliography}

\end{document}